\documentclass[preprint,12pt]{elsarticle}

\usepackage{graphics}
\usepackage{verbatim}

\usepackage{amssymb}
\usepackage{amsthm}
\usepackage{amsmath}

\usepackage{algpseudocode}
\usepackage{algorithm}

\usepackage{hyperref}
 \usepackage{listings}
  \usepackage{courier}
  \usepackage{caption}
\DeclareCaptionFont{white}{\color{white}}
\DeclareCaptionFormat{listing}{\colorbox[cmyk]{0.43, 0.35, 0.35,0.01}{\parbox{\textwidth}{\hspace{15pt}#1#2#3}}}
\journal{Information Sciences}

\begin{document}

\begin{frontmatter}



\title{CACO : Competitive Ant Colony Optimization, A Nature-Inspired Metaheuristic For Large-Scale Global Optimization}

\author[label1]{M. A. El-Dosuky}
\address[label1]{Computer sciences Department, Faculty of Computers and Information, P.D.Box 35516, Mansoura University, Egypt}

\begin{abstract}
Large-scale problems are nonlinear problems that need metaheuristics, or global optimization algorithms. This paper reviews nature-inspired metaheuristics, then it introduces a framework named Competitive Ant Colony Optimization inspired by the chemical communications among insects. Then a case study is presented to investigate the proposed framework for large-scale global optimization.
\end{abstract}


\end{frontmatter}


\section{Introduction}
Large-scale problems are nonlinear problems that need metaheuristics, or global optimization algorithms (\cite{Deb1995},\cite{Glover1986}).

This paper reviews nature-inspired metaheuristics, then it introduces a framework named Competitive Ant Colony Optimization inspired by the chemical communications among insects. Then a case study is presented to investigate the proposed framework. Finally, the paper concludes with a discussion of future works.

\section{Nature-Inspired Metaheuristics}
Large-scale problems are nonlinear problems that need metaheuristics, or global optimization algorithms (\cite{Deb1995},\cite{Glover1986}).
Many nature-inspired metaheuristic optimization algorithms are proposed to imitate the best behaviors in nature \cite{Yang2008} such as the artificial immune system \cite{Farmer1986}, genetic algorithms \cite{Goldberg1989}, ant colony optimization (ACO) \cite{Dorigo1992}, particle swarm optimization (PSO) \cite{Kennedy1995}, Artificial Bee Colony Algorithm (ABC) \cite{Karaboga2005}, and cuckoo search \cite{Yang2009}. Recently, many nature inspired algorithms are proposed(\cite{Eldosuky1}, \cite{Eldosuky2}, \cite{Eldosuky3}, \cite{Eldosuky4}).
A metaheuristic explores the search space by employing two components of intensification and diversification (\cite{Zapfel2010}, \cite{Gazi2004}).  Intensification strategy focuses on examining neighbors of elite solutions while diversification strategy encourages examining unvisited regions \cite{Glover1997}. Intensification is a deterministic component and diversification is a stochastic component \cite{Hoos2005}. Metaheuristic algorithms should be designed so that intensification and diversification play balanced roles \cite{Blum2003}.

\subsection{Ant colony optimization (ACO)}
Ant colony optimization (ACO) generates artificial ants that move on the problem graph depositing artificial pheromone so that the future artificial ants can build better solutions (\cite{Dorigo1992}, \cite{Dorigo2005}). ACO has been successfully applied to an impressive number of optimization problems especially for routing and scheduling problems (\cite{Santos2010}, \cite{Bell2004}).

ACO proves reliability in large-scale applications such as large-distorted fingerprint matching \cite{Cao2012} and solving the logistics problem arising in disaster relief activities\cite{Yi2007}.

\section{Competitive Ant Colony Optimization}
It is probably safe to say that insects rely more heavily on chemical signals than on any other form of communication.  These signals, often called semiochemicals or infochemicals, serve as a form of language that helps to mediate interactions between organisms.  Insects may be highly sensitive to low concentrations of these chemicals in some cases, a few molecules may be enough to elicit a response.
Semiochemicals can be divided into Pheromones and Allelochemicals based on who sends a message and who receives it \cite{chemcomm}.
 Pheromones are chemical signals that carry information from one individual to another member of the same species.  These include sex attractants, trail marking compounds, alarm substances, and many other intraspecific messages.
Allelochemicals are signals that travel from one animal to some member of a different species.  These include defensive signals such as repellents, compounds used to locate suitable host plants, and a vast array of other substances that regulate interspecific behaviors.

Allelochemicals can be further subdivided into three groups based on who benefits from the message:
\begin{description}
  \item[Allomones]  benefit the sender such as a repellent, or defensive compound (e.g. cyanide) that deters predation..
  \item[Kairomones]  benefit the receiver -- such as an odor that a parasite uses to find its host.
  \item[Synomones]  benefit both sender and receiver -- such as plant volatiles that attract insect pollinators..
\end{description}

The diffusion equation of chemical signals is defined as\cite{Happ1974}:
 
$K= \frac{Q}{2D\pi r}\mathit{efrec}\frac{r}{\sqrt{4Dt}}$

where Q, D, and K are emission rate, diffusion coefficient, and threshold concentration, respectively, and where r is the "radius" of the active space (cm), t is the time from the beginning of emission, and where efrc(x) is the complementary error function.

\subsection{Proposed Metaheuristic}
In their search for food, ants use pheromones to communicate. Assuming there is a natural battle between ants and their enemy that produces allomones. The enemy can be other incest species or ants of different kind. Let assume that the enemy is a group of wasps.
Based upon these assumptions, let us propose the following scenario.

First, Ants and Wasps are two groups of insects, competing in the same environment to search for food. Each group behaves like traditional ACO algorithm.
Second, within the same group, communication is done using pheromones. Communication between the two groups is done using allomones.
Third, Wasps can kill ants if they are close enough. 
 
\subsection{Implementation}
Implementation of this modification is done in Microsoft Visual C\# . The code listing is shown below.
\begin{lstlisting}
    class Wasp : ACO.ACO
    {
        public Wasp(Dataset data, float evapore, float aging, float limit, bool useOptimize)
            : base(data, evapore, aging, limit, useOptimize)
        {
        }
        public override Graph Optimize(params float[] parameters)
        {
            return this.ThisGraph;
        }
    }

    class Ant : ACO.ACO
    {
        Wasp wasps;
        public Ant(Wasp wasps, Dataset data, float evapore, float aging, float limit, bool useOptimize)
            : base(data, evapore, aging, limit, useOptimize)
        {

            this.Optimize(1, 1, 1);

            this.wasps = wasps;

        }
        public override Graph Optimize(params float[] parameters)
        {
            Graph g = new Graph(this);


            float r = parameters[0];
            float d = parameters[1];
            float q = parameters[2];
            int k;

            double sum = 0;
            int iterations = 1000;
            for (int t = 1; t < iterations; t++)
            {
                sum = q / (2*Math.PI*r) * r / (Math.Sqrt(4*d*t));

                k =(int) sum * iterations;

               efrc( new Graph(this.wasps) , this.ThisGraph);
            }



            return this.ThisGraph;
        }

        public void efrc(Graph w, Graph a)
        {
            w.Complement(a);
        }
    }


    class Program
    {
        static void Main(string[] args)
        {
            Dataset d1 = Dataset.Load(Datasets.Audiology);
            Dataset d2 = Dataset.Load(Datasets.BreastCancer);
            Dataset d3 = Dataset.Load(Datasets.Mushroom);
            Dataset d4 = Dataset.Load(Datasets.Vote);
            Dataset d5 = Dataset.Load(Datasets.Wine);

            work(d1);
            work(d2);
            work(d3);
            work(d4);
            work(d5);

            Console.ReadKey();

        }
        static void work(Dataset dataset)
        {
            Wasp wasps = new Wasp(dataset, 0.2F, 1.0F, 1.0F, false);
            Ant ants = new Ant(wasps, dataset, 0.2F, 1.0F, 1.0F, true);
            
            Console.WriteLine(string.Join("\t", new string[] 
                { 
                dataset.name , 
                dataset.size.ToString(), 
                dataset.feats.ToString(), 
                wasps.feats.ToString(), 
                ants.feats.ToString() }));
        }
    }
\end{lstlisting}
\section{Evaluation}
Experiments are carried out on  five datasets which are all from UCI datasets (\url{http://archive.ics.uci.edu/ml/datasets.html}). In order to find whether our algorithm could find an optimal reduct, we compare algorithm of with traditional method . The experiments are summarized in Table 1.

\begin{table}[ht]
\caption{Comparison}
\begin{center}
    \begin{tabular}{ | l | l | l | l | p{4cm} |}
    \hline
    Dataset & Instants & Features &	ACO	& Proposed CACO \\ \hline
    Audiology & 200 & 70 & 20 & 12\\ \hline
    Breast Cancer & 699 & 10 & 4 & 4\\ \hline
    Mushroom & 8124 & 23 & 6 & 5\\ \hline
    Wine & 178 & 14 & 6 & 5\\ \hline
    Vote & 435 & 17 & 12 & 10\\ \hline
    \end{tabular}
\end{center}
\end{table}

\section{Conclusion}
This paper reviews Ant Colony optimization algorithms and describes a new heuristic optimization method based on swarm intelligence. It presents a mechanism for enhancing Ant colony optimization by introducing the natural battle between ants and their enemy that produces Allomones.
It is very simple, easily implemented and it needs fewer parameters, which made it fully developed and applied for feature extraction task

%





\bibliographystyle{plainnat}



\end{document}